\documentclass[letterpaper, 10 pt, conference]{ieeeconf}  

\IEEEoverridecommandlockouts                              

\usepackage{graphicx}
\usepackage{graphics} 
\usepackage{amssymb}  
\usepackage{amsmath}
\usepackage{url}
\usepackage{booktabs}
\usepackage{multirow}    
\usepackage{makecell}    
\usepackage{adjustbox}   
\usepackage{pifont}      
\usepackage[justification=raggedright,singlelinecheck=false]{caption}

\usepackage{xcolor}

\title{\LARGE \bf
How Should a Robot Configure Its Laser Scanner for Inspection?
}

\author{Zhiling Chen$^{1}$, David Gorsich$^{2}$, Matthew P. Castanier$^{2}$, Yang Zhang$^{1}$, Jiong Tang$^{1}$, and Farhad Imani$^{1}$%
\thanks{*This work was supported under Cooperative Agreement W56HZV-21-2-0001 with the US Army DEVCOM Ground Vehicle Systems Center (GVSC), through the Virtual Prototyping of Autonomy Enabled Ground Systems (VIPR-GS) program, and by the National Science Foundation, United States (Grant No. 2434519).}%
\thanks{**DISTRIBUTION STATEMENT A. Approved for public release; distribution is unlimited. OPSEC10399}%
\thanks{$^{1}$Zhiling Chen, Yang Zhang, Jiong Tang, and Farhad Imani are with the University of Connecticut, Storrs, CT, USA.}%
\thanks{$^{2}$David Gorsich and Matthew P. Castanier are with the US Army DEVCOM Ground Vehicle Systems Center (GVSC), Warren, MI, USA.}%
}

\begin{document}

\maketitle
\thispagestyle{empty}
\pagestyle{empty}



\begin{abstract}

Robotic inspection relies on accurate sensing to acquire high-fidelity geometric measurements for defect detection and metrology. While prior work has focused on robot motion and viewpoint planning, how to configure sensing parameters remains largely underexplored, despite their decisive impact on measurement quality. We propose SenseHD, a robotic sensing system that formulates scanner configuration as an instruction-conditioned sensing decision. Instead of predicting precise parameter values, SenseHD treats sensing parameters as discrete sensing actions and selects stable sensing regimes through hyperdimensional associative memory. Experiments on a real robotic inspection platform demonstrate that SenseHD robustly selects appropriate sensing configurations and significantly improves inspection reliability, while remaining lightweight and efficient compared to baseline methods.

\end{abstract}

\section{INTRODUCTION}
\label{introduction}

Robotic inspection systems are widely used in manufacturing, remanufacturing, and quality assurance to acquire geometric measurements for defect detection, dimensional verification, and metrology \cite{papavasileiou2025quality}.
In many deployments, inspection pipelines rely on predefined scanning trajectories and fixed (or manually tuned) sensing configurations that are repeatedly applied across similar parts and environments \cite{almadhoun2016survey}.
Such designs work well in structured settings, but they implicitly assume stable sensing conditions and consistent inspection requirements.
In practice, these assumptions often break: variations in object geometry, surface reflectivity, lighting, and inspection intent can significantly alter sensing outcomes, making fixed configurations brittle and hard to generalize.
These challenges motivate an embodied inspection perspective, where sensing decisions are coupled with task specification and physical interaction rather than treated as static parameters \cite{bajcsy1988active,bajcsy2018revisiting,chen2025scanbot}.

As illustrated in Fig.~\ref{introduction}, embodied inspection can be viewed as a closed-loop process spanning task initialization, sensing configuration, physical scanning, and measurement-driven analysis. 
Within this process, sensing configuration directly shapes the quality and completeness of acquired measurements. 
Despite substantial progress in robot motion planning and inspection trajectory generation, how to configure sensing parameters in a task- and context-aware manner remains largely underexplored.
Fig.~\ref{introduction} further illustrates this challenge through representative inspection outcomes. 
Given the same inspection instructions for an object, unsuitable scanner configurations can lead to inspection failure: an overly narrow measurement range results in incomplete coverage (Fig.~\ref{introduction}a), while insufficient exposure or illumination degrades fine details (Fig.~\ref{introduction}b). 
In contrast, selecting a configuration that balances coverage, resolution, and signal quality enables successful inspection (Fig.~\ref{introduction}c) \cite{shin2019camera}. 
These examples underscore that scanner configuration is an integral component of the inspection task rather than a secondary system setting.

\begin{figure}[t]
  \centering
  \includegraphics[width=0.5\textwidth]{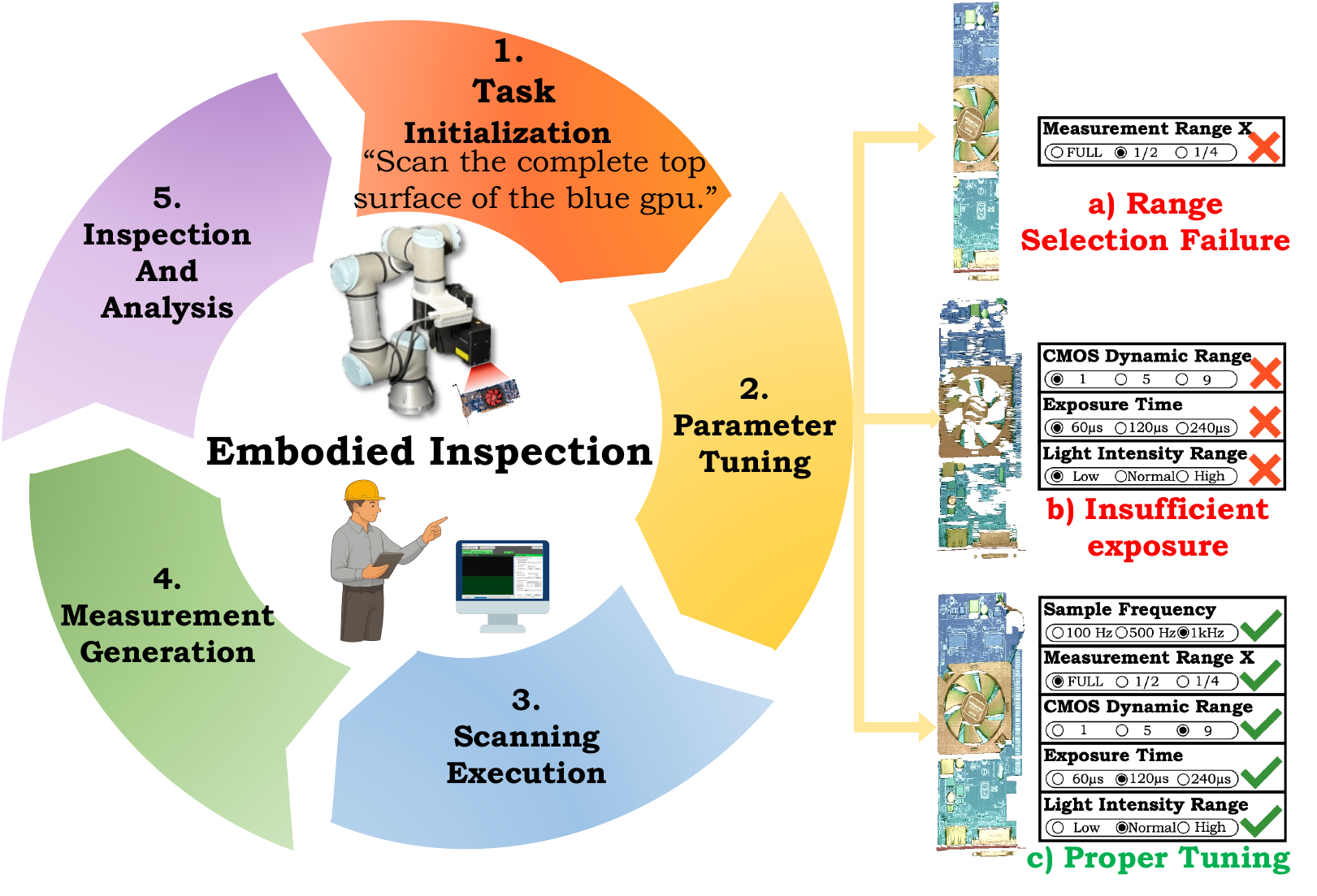} 
  \caption{Embodied inspection process and motivation for instruction-conditioned sensing. }
  \label{introduction}
\end{figure}


These observations highlight that selecting sensing parameters is a challenging decision problem.
The choice depends on multiple factors, including object geometry, surface reflectivity, illumination conditions, and inspection intent.
Existing approaches often rely on expert manual tuning or cast parameter selection as continuous regression.
However, predicting precise values is brittle and hard to generalize under appearance variation and environmental shift.
Moreover, collecting large-scale labeled data to supervise precise parameter regression is costly, making such approaches less suitable for data-scarce settings and cold-start deployment on new parts, materials, or sensing setups.


In this work, we take a different perspective. Instead of predicting exact parameters, we cast scanner configuration as selecting a stable sensing regime analogous to invoking pre-validated `recipes' on industrial controllers. Given an instruction and a pre-scan observation, the robot retrieves an appropriate sensing action from memory. This regime-based view prioritizes reliability and direct compatibility with standard Manufacturing Execution Systems (MES) \cite{mantravadi2019overview}, and is data-efficient: regimes can be defined with few exemplars and expanded by adding new experiences, enabling cold-start deployment without retraining.

To support such associative sensing decisions, the underlying representation must be robust to appearance variation, enable retrieval over discrete actions, and remain lightweight for real-time deployment.
It should also admit online adaptation: as the sensing conditions drift, the robot should update its memory incrementally on-device.
Hyperdimensional Computing (HDC) naturally satisfies these requirements by encoding observations into high-dimensional distributed vectors (hypervectors) and storing them in a lightweight associative memory that can be updated incrementally on-device~\cite{ge2020classification}. At inference time, we form a query hypervector and retrieve the closest stored prototype(s) using a similarity metric (e.g., cosine similarity). The retrieved exemplars and their similarity scores provide a transparent, interpretable, and traceable rationale for the resulting sensing configuration.
Building on this idea, we propose SenseHD, a robotic sensing system that performs instruction-conditioned sensing decisions using hyperdimensional representations.
SenseHD jointly encodes visual observations and inspection instructions and retrieves scanner configurations through associative memory.

We implement SenseHD on a real robotic inspection platform equipped with an industrial laser profiler and evaluate it across diverse objects and inspection tasks. Experimental results demonstrate that SenseHD reliably selects appropriate sensing configurations, generalizes across appearance variations, and significantly improves inspection robustness compared to baseline approaches.

In summary, this paper makes the following contributions:
\begin{itemize}
    \item We identify scanner configuration as a first-class sensing action in robotic inspection and formalize instruction-conditioned sensing as a decision problem.
    \item We introduce SenseHD, a memory-based sensing system that uses hyperdimensional representations for robust and efficient scanner parameter selection.
    \item We validate SenseHD on a real robotic platform and show its effectiveness across diverse objects, inspection intents, and appearance variations.
\end{itemize}

\section{Related Work}




HDC, also known as Vector Symbolic Architecture (VSA), represents information using high-dimensional distributed vectors, referred to as hypervectors, and performs inference through similarity-based comparison \cite{kanerva2009hyperdimensional}. By encoding sensory inputs or symbolic variables into a common hyperdimensional space and retrieving the closest matching prototype, HDC realizes associative memory that is computationally efficient and robust to noise, making it well suited for robotic applications \cite{kwon2024brain}. Prior work has applied HDC primarily to classification \cite{11246660} and control tasks \cite{ni2022hdpg}, where actions are selected by associating observed states with stored prototypes. However, these approaches typically focus on perception or control under simulated or highly controlled settings and assume fixed sensing configurations. In contrast, this paper adopts a sensing-centric perspective for real-world robotic inspection, treating sensing parameters as discrete sensing actions. By casting parameter selection as nearest-prototype retrieval in a shared hyperdimensional space, we reduce the decision to similarity search over discrete sensing recipes, enabling a principled and reproducible mapping from observations to controller-valid parameter configurations under appearance variation.

\section{Robotic Inspection Problem and System Setup}

\subsection{Problem Formulation}
\label{sec:problem_definition}

Robotic inspection differs from conventional perception in that sensing itself is a controllable robotic action rather than a passive observation step. In laser-based inspection, measurement quality depends not only on robot motion but also critically on the sensor configuration chosen prior to data acquisition. Although rescanning is possible in principle, it is often impractical under production cycle-time and repeatability constraints: a suboptimal configuration can saturate or truncate the signal, yielding scans that are effectively unusable and costly to redo.

We consider an inspection scenario in which a manipulator-mounted laser profiler performs a scan after a short pre-sensing phase. At configuration time, the robot observes a context $c=(o,x)$, where $o$ is a pre-scan visual observation (e.g., an RGB image) and $x$ is a natural-language instruction specifying the inspection intent and required level of detail.

Before acquiring any laser data, the robot must choose a sensing configuration $\theta\in\Theta$. Each configuration is a combination of $K$ discrete sensing parameters, $\theta=(\theta_1,\ldots,\theta_K)$ with $\theta_k\in\Theta_k$, yielding a combinatorial action space $\Theta=\Theta_1\times\cdots\times\Theta_K$. Executing $\theta$ determines the physical sensing regime and thus scan fidelity.

\noindent\textbf{Budgeted formulation.}
In principle, the robot could adaptively try multiple configurations within a limited trial budget $B$ (e.g., due to cycle-time constraints). Let $\theta_{1:B}=\{\theta_1,\ldots,\theta_B\}$ denote a sequence of attempted configurations. We formulate budgeted configuration selection as:
\begin{equation}
\max_{\theta_{1:B}\subseteq \Theta}\ \mathbb{E}\!\left[\max_{t\le B} U(\theta_t; o,x)\;-\;\sum_{t=1}^{B}\lambda\,C(\theta_t)\right],
\end{equation}
where $U(\theta;o,x)$ measures scan utility (e.g., geometric fidelity / defect resolvability) and $C(\theta)$ captures the cost of an attempt (time, wear, risk), with trade-off $\lambda$. In industrial inspection, $B$ is typically very small and often effectively $B=1$, making one-shot selection the practically relevant regime. We therefore focus on learning a one-shot predictor $f_\phi:\mathcal{O}\times\mathcal{X}\rightarrow\Theta$ that outputs a near-optimal configuration prior to scan execution:
\begin{equation}
\theta \;=\; f_\phi(o,x).
\end{equation}

\subsection{Robotic Inspection Platform}
\label{sec:platform}

\begin{figure}[t]
  \centering
  \includegraphics[width=0.45\textwidth]{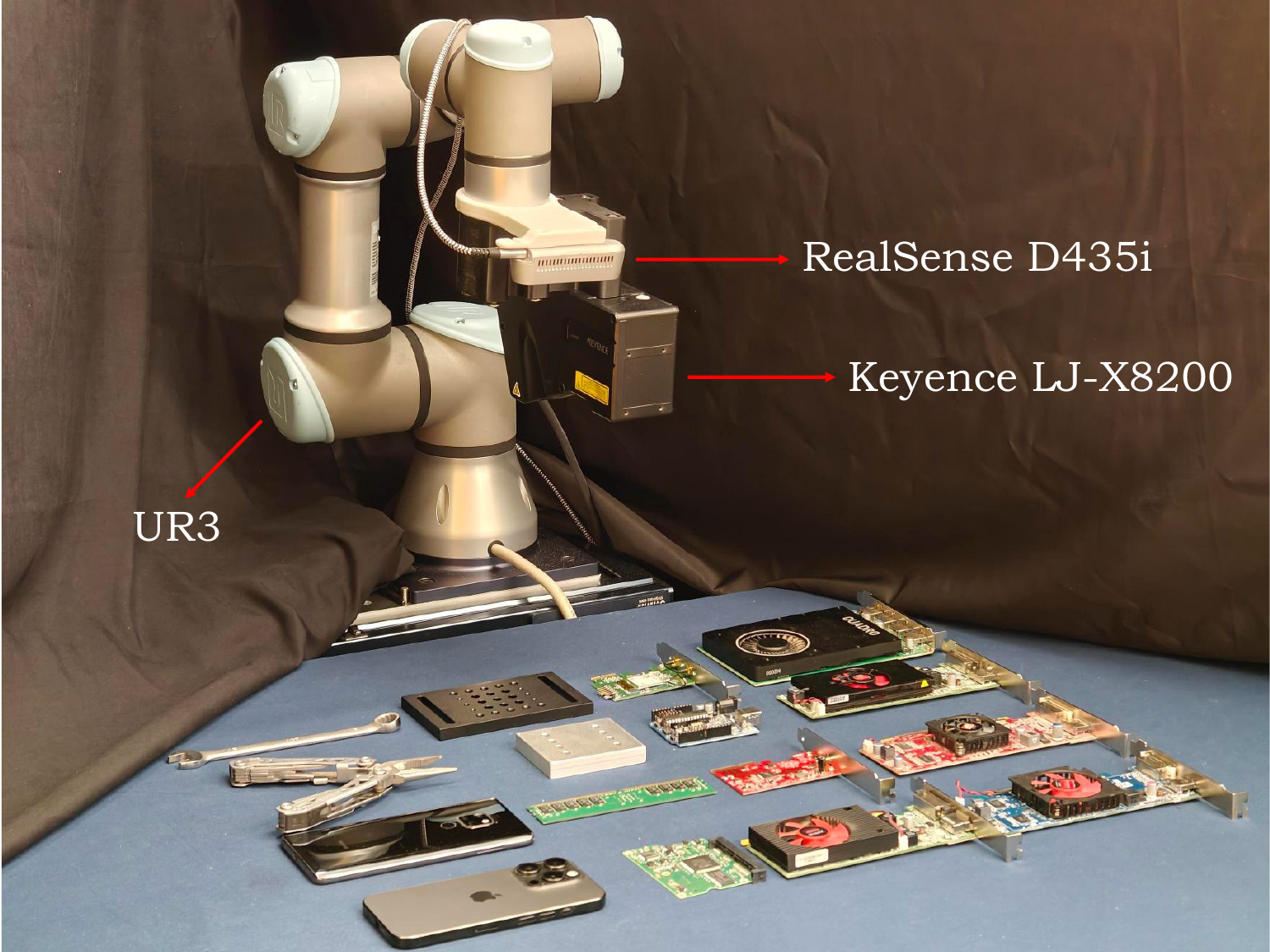} 
  \caption{Robotic inspection platform used in this work. A UR3 robotic arm is equipped with a Keyence LJ-X8200 laser profiler for inspection and a RealSense D435i RGB-D camera for pre-scan observation.}
  \label{setup}
\end{figure}

The robotic inspection platform used in this work is shown in Fig.~\ref{setup}. The system consists of a 6-DOF collaborative robotic manipulator equipped with an industrial laser profiler and an RGB-D camera. A UR3 robotic arm positions and moves the sensing payload during inspection, with a Keyence LJ-X8200 laser displacement sensor rigidly mounted on the end-effector as the primary measurement device. The laser profiler is interfaced with an LJX-8000A controller for data acquisition and parameter configuration. The laser profiler acquires high-resolution 2D surface profiles and exposes multiple configurable parameters that are set programmatically prior to scan execution. To support instruction-conditioned sensing decisions, an Intel RealSense D435i RGB-D camera is co-mounted near the laser profiler and used exclusively in the pre-sensing phase. The captured RGB observations provide appearance cues such as geometry, scale, surface reflectivity, and illumination conditions.

\subsection{Data Collection}

\begin{figure}[t]
  \centering
  \includegraphics[width=0.5\textwidth]{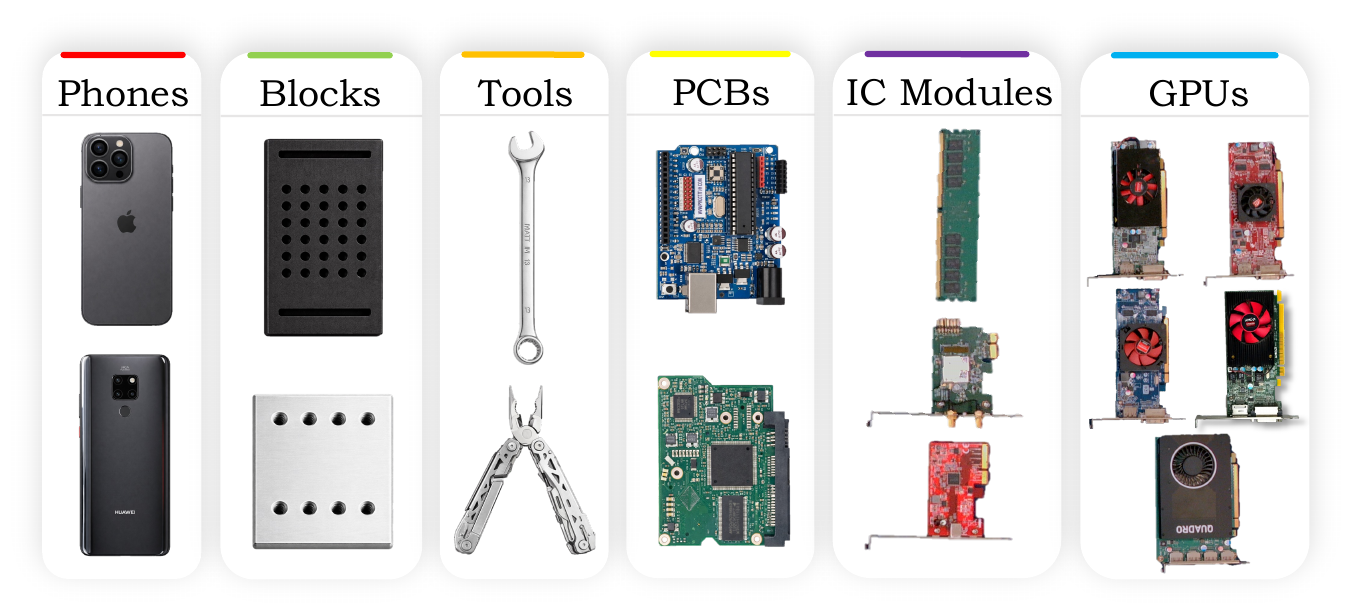} 
  \caption{Inspection objects used in our experiments. The experimental set consists of six categories of industrial objects: Phones, PCBs, GPUs, Blocks, Tools, and IC Modules.}
  \label{fig:objects}
\end{figure}


To support instruction-conditioned sensing decisions on a real robotic platform, we collect inspection data using the physical system described in Section~\ref{sec:platform}. The dataset is grounded in a structured taxonomy of \emph{industrial quality inspection tasks}, spanning appearance inspection (global/local, outline/detail) and geometric metrology (dimensional verification, registration). This taxonomy provides a principled basis for designing semantically consistent inspection instructions. The dataset covers a diverse set of 16 industrial objects with substantial variation in geometry and material properties, including consumer electronics, printed circuit boards (PCBs), GPU modules, mechanical tools, and IC modules, as shown in Fig.~\ref{fig:objects}.

For each inspection instance, the robot acquires a pre-scan RGB observation, receives a task-conditioned instruction, and executes a specific scanner configuration. 
Consistent with the Keyence LJ-X8000 controller paradigm, each of the five sensing parameters is chosen from a finite, controller-defined discrete option set (with parameter-specific cardinalities). To avoid subjective micro-tuning while spanning valid operating regimes, we select three representative controller-valid levels per parameter (lower / mid / upper on the controller’s numeric scale), determined via pilot scans under objective capture criteria (e.g., non-saturated returns and stable profile acquisition). We treat each resulting pre-validated 5D configuration vector as a sensing recipe.
This formulation mirrors the industrial practice of utilizing pre-validated configuration profiles over manual micro-tuning, ensuring the learned policy is directly compatible with standard MES. Furthermore, by decoupling decision logic from specific hardware constraints, this design enables the policy to generalize across different scanner models via simple parameter remapping.

As illustrated by the \ding{182} \emph{Data Evolution Flywheel} in Fig.~\ref{fig:sensehd_overview}, data collection follows an iterative process to ensure physical validity. We employ GPT-5-based knowledge distillation to generate diverse task-conditioned instructions, while expert-guided calibration filters out infeasible or unstable configurations. This calibrated process produces high-quality instruction--observation--action instances, ensuring both linguistic diversity and task-appropriateness. The final dataset contains 856 instances across all objects and tasks.


\section{SenseHD: Instruction-Conditioned Sensing System}

\begin{figure*}[t]
  \centering
  \includegraphics[width=\textwidth]{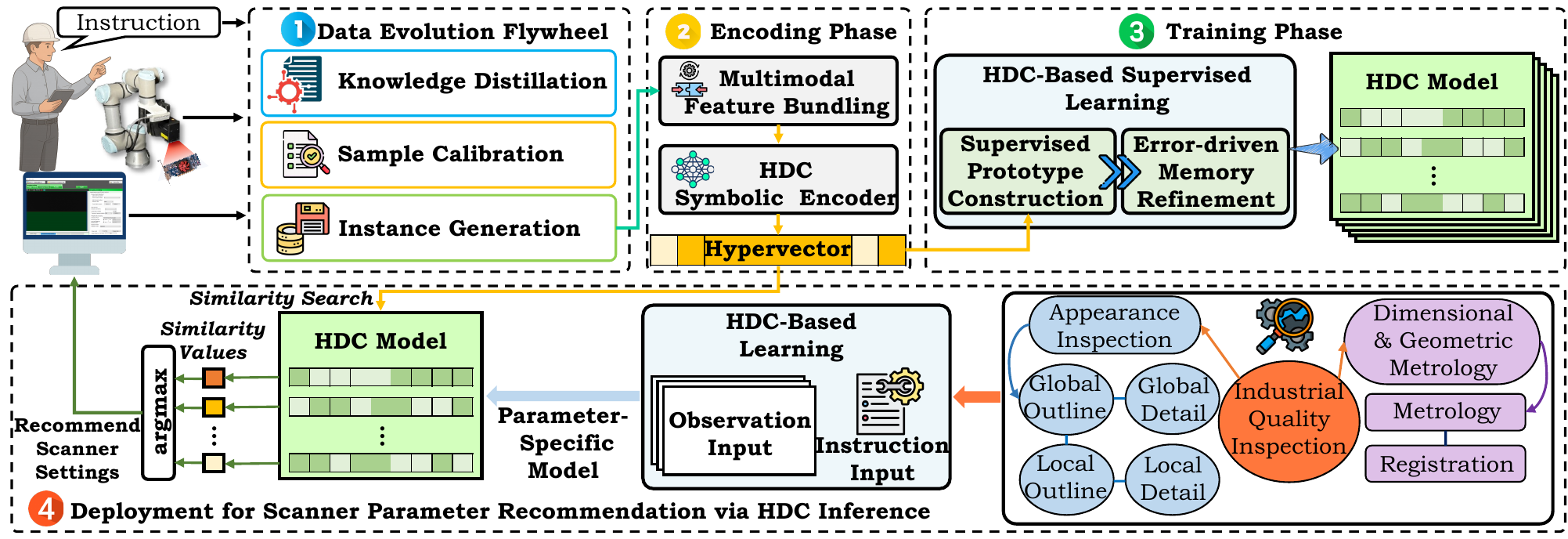} 
  \caption{Overview of SenseHD.}
  \label{fig:sensehd_overview}
\end{figure*}

In this paper, we present \textbf{SenseHD}, an HDC-based learning framework for instruction-conditioned scanner parameter recommendation in industrial inspection.
As shown in Fig.~\ref{fig:sensehd_overview}, SenseHD begins by organizing raw sensing data and expert knowledge through a data evolution flywheel (\ding{182}), which consolidates domain knowledge, calibrates representative scanning samples, and produces structured instruction--observation instances suitable for learning.
Based on these instances, visual observations and natural-language instructions are jointly encoded into symbolic hypervectors via multimodal feature bundling and hyperdimensional encoding (\ding{183}), yielding a compact representation that preserves both semantic intent and appearance cues. Learning in SenseHD is formulated as a supervised HDC problem, where parameter-specific associative memories are constructed by aggregating encoded hypervectors according to their corresponding scanner settings (\ding{184}). Once trained, the resulting HDC models support fast similarity-based inference at deployment time (\ding{185}), allowing SenseHD to directly retrieve compatible scanner parameter configurations for unseen instruction--observation inputs.

\subsection{Encoding Phase}
\label{sec:data_encoding}

\begin{figure}[t]
  \centering
  \includegraphics[width=0.5\textwidth]{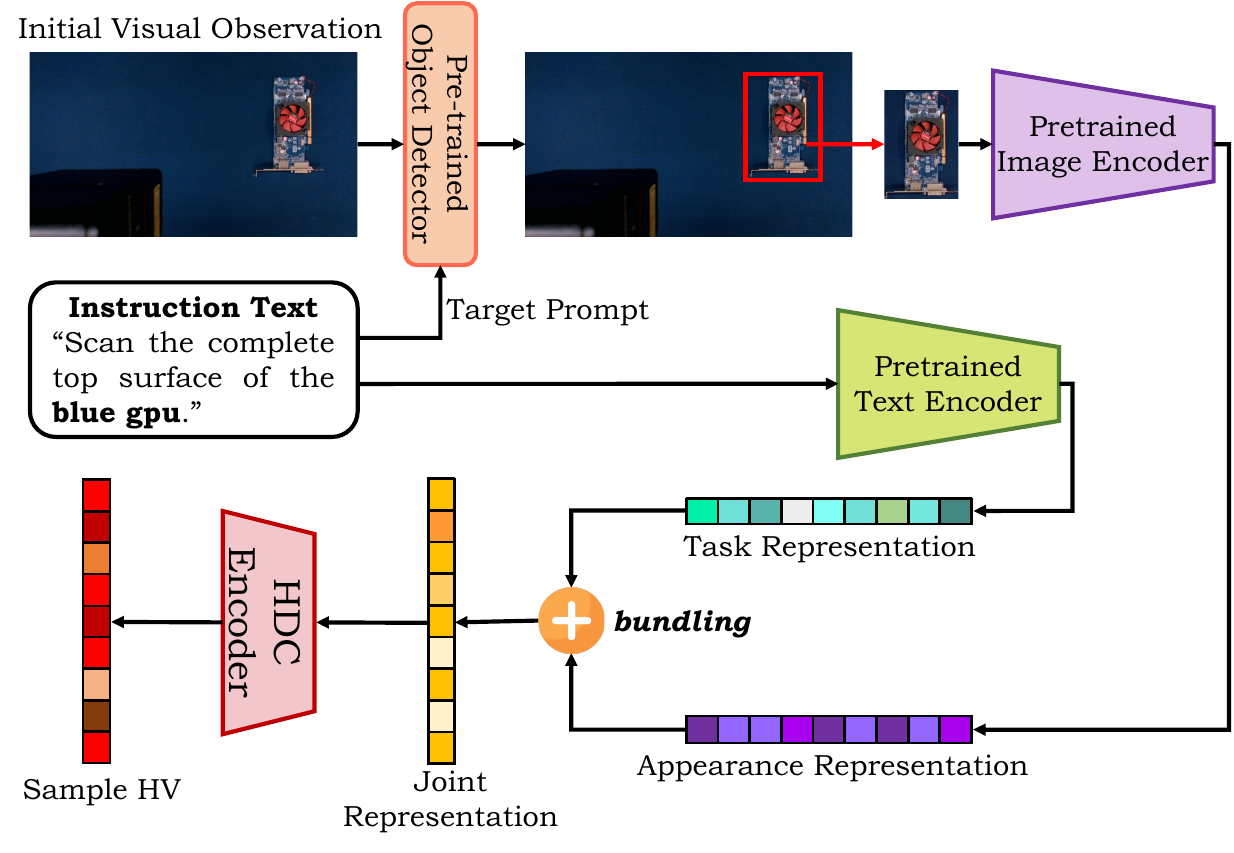} 
  \caption{Encoding Phase of SenseHD.}
  \label{fig:data_encoding}
\end{figure}

Given an instruction--observation pair, SenseHD encodes multimodal inputs into a unified hyperdimensional representation suitable for associative learning.
As illustrated in Fig.~\ref{fig:data_encoding}, the encoding process starts from an input image $\mathbf{I} \in \mathcal{I}$ and a natural-language instruction $\mathbf{T}$.

\noindent\textbf{Prompt-conditioned Object Localization.}
To focus on task-relevant regions and suppress background interference, SenseHD first applies a pre-trained object detection model (e.g., YOLO-World \cite{cheng2024yolo})
$f_{\text{det}}: \mathcal{I} \rightarrow \mathcal{B}$,
where $\mathcal{B}$ denotes a set of bounding boxes.
The detector is guided by a target prompt derived from the instruction text, enabling prompt-conditioned localization of the object of interest.
Given the detected bounding box $\mathbf{b} \in \mathcal{B}$, a region of interest (ROI) image $\mathbf{I}_{\text{roi}} \subset \mathbf{I}$ is obtained and used for subsequent encoding.

\noindent\textbf{Multimodal Feature Extraction.}
We adopt a pre-trained vision--language model, \emph{e.g.}, OpenAI CLIP~\cite{radford2021learning}, to extract paired image and text embeddings without task-specific fine-tuning. 
Given a cropped ROI image $\mathbf{I}_{\text{roi}}$, the image encoder produces an embedding $\mathbf{e}_{\text{img}} = f_{\text{img}}(\mathbf{I}_{\text{roi}}) \in \mathbb{R}^{d}$, while the instruction text $\mathbf{T}$ is mapped by the text encoder to $\mathbf{e}_{\text{txt}} = f_{\text{txt}}(\mathbf{T}) \in \mathbb{R}^{d}$. 
Both encoders operate in a shared embedding space of dimension $d$, where $\mathbf{e}_{\text{img}}$ captures the visual appearance of the target object and $\mathbf{e}_{\text{txt}}$ encodes the inspection intent specified by the instruction.

\noindent\textbf{Joint Representation and HDC Projection.}
To integrate appearance and task semantics, SenseHD combines the two embeddings using a bundling operation,
\begin{equation}
\mathbf{e}_{\text{joint}} = \mathcal{B}(\mathbf{e}_{\text{img}}, \mathbf{e}_{\text{txt}}).
\end{equation}
The joint embedding is then mapped into a high-dimensional symbolic space using an HDC encoder
$\Phi: \mathbb{R}^{d} \rightarrow \mathbb{R}^{D}$,
\begin{equation}
\mathbf{h} = \Phi(\mathbf{e}_{\text{joint}}), \qquad \mathbf{h} \in \mathbb{R}^{D},
\end{equation}
where $D \gg d$.
The resulting hypervector $\mathbf{h}$ serves as the sample-level representation used for subsequent HDC-based supervised learning and inference.

\begin{table*}[t]
\centering
\caption{\textbf{Main Results on our Dataset.}
Each model predicts 5 scanning parameters. 
For each parameter we report Exact Accuracy, Win@1 Accuracy, and F1.}
\label{tab:main_results}
\resizebox{\textwidth}{!}{
\begin{tabular}{l|ccc|ccc|ccc|ccc|ccc|ccc|ccc}
\toprule
& \multicolumn{3}{c|}{\textbf{Sample Freq}} 
& \multicolumn{3}{c|}{\textbf{Measurement Range X}} 
& \multicolumn{3}{c|}{\textbf{Exposure Time}} 
& \multicolumn{3}{c|}{\textbf{CMOS Dynamic Range}} 
& \multicolumn{3}{c|}{\textbf{Control Light Intensity}} 
& \multicolumn{3}{c}{\textbf{Average}} \\
\textbf{Method} 
& Exact & Win@1 & F1
& Exact & Win@1 & F1
& Exact & Win@1 & F1
& Exact & Win@1 & F1
& Exact & Win@1 & F1
& Exact & Win@1 & F1 \\ 
\midrule
\multicolumn{19}{l}{\textbf{A. Rule-based Baselines}} \\
Rule-based Heuristic 
& 39.9$\pm$1.8 & 59.1$\pm$2.6 & 23.3$\pm$1.7 
& 56.5$\pm$4.7 & - & 24.0$\pm$1.3 
& 47.7$\pm$2.1 & 78.4$\pm$2.7 & 21.9$\pm$1.2 
& 53.1$\pm$3.0 & 89.2$\pm$2.4 & 25.3$\pm$0.9 
& 54.7$\pm$4.7 & \textbf{100.0$\pm$0.0} & 23.8$\pm$1.7 
& 50.4$\pm$3.3 & 81.7$\pm$1.7 & 23.7$\pm$1.4\\
\midrule

\multicolumn{19}{l}{\textbf{B. Instruction-Only Methods}} \\
Logistic Regression
& \textbf{97.9$\pm$1.4} & \textbf{98.6$\pm$0.9} & \textbf{98.0$\pm$1.4} 
& 93.1$\pm$1.7 & - & 91.8$\pm$1.7 
& 75.3$\pm$2.3 & 92.7$\pm$1.3 & 74.1$\pm$2.3 
& 83.3$\pm$2.1 & 96.5$\pm$0.5 & 81.5$\pm$2.5 
& 90.2$\pm$0.7 & 95.5$\pm$1.1 & 88.3$\pm$1.2 
& 88.0$\pm$1.6 & 95.8$\pm$0.5 & 86.7$\pm$1.8 \\

KNN
& 92.6$\pm$2.3 & 95.7$\pm$2.4 & 92.6$\pm$2.0 
& 83.4$\pm$1.2 & - & 82.1$\pm$2.1 
& 74.8$\pm$1.5 & 91.9$\pm$1.6 & 73.3$\pm$1.3 
& 80.9$\pm$2.6 & 95.5$\pm$1.3 & 77.7$\pm$3.8 
& 79.3$\pm$1.4 & 93.5$\pm$1.2 & 75.2$\pm$0.8 
& 82.2$\pm$1.8 & 94.1$\pm$1.6 & 80.2$\pm$2.0 \\

\midrule

\multicolumn{19}{l}{\textbf{C. Observation-Only Methods}} \\
ResNet
& 42.2$\pm$3.6 & 62.6$\pm$3.6 & 28.5$\pm$3.4 
& 71.2$\pm$2.1 & - & 70.5$\pm$2.2 
& 74.8$\pm$2.4 & \textbf{100.0$\pm$0.0} & 73.7$\pm$3.2 
& 73.7$\pm$3.1 & \textbf{100.0$\pm$0.0} & 74.8$\pm$4.8 
& \textbf{100.0$\pm$0.0} & \textbf{100.0$\pm$0.0} & \textbf{100.0$\pm$0.0} 
& 72.4$\pm$2.2 & 90.7$\pm$0.9 & 69.5$\pm$2.7 \\

ViT
& 44.2$\pm$3.1 & 65.0$\pm$3.7 & 31.3$\pm$4.5 
& 71.4$\pm$1.7 & - & 72.2$\pm$2.8 
& 72.2$\pm$2.9 & \textbf{100.0$\pm$0.0} & 72.1$\pm$2.4 
& 74.1$\pm$2.7 & \textbf{100.0$\pm$0.0} & 77.0$\pm$2.1 
& \textbf{100.0$\pm$0.0} & \textbf{100.0$\pm$0.0} & \textbf{100.0$\pm$0.0} 
& 72.4$\pm$2.1 & 91.3$\pm$0.9 & 70.5$\pm$2.4 \\

\midrule

\multicolumn{19}{l}{\textbf{D. Fusion Methods}} \\
DNN
& 85.1$\pm$3.8 & 90.9$\pm$0.9 & 84.2$\pm$6.2 
& 91.2$\pm$4.3 & - & 91.6$\pm$4.5 
& 84.5$\pm$2.8 & \textbf{100.0$\pm$0.0} & 85.2$\pm$2.8 
& \textbf{86.2$\pm$0.8} & 99.8$\pm$0.5 & \textbf{85.6$\pm$1.3} 
& 98.8$\pm$0.6 & 99.4$\pm$0.4 & 98.6$\pm$0.7 
& 89.2$\pm$2.5 & 97.5$\pm$0.3 & 89.0$\pm$3.1\\


\midrule

\multicolumn{19}{l}{\textbf{E. Multimodal Large Language Models}} \\
Qwen3-VL-4B-Instruct
& 60.0$\pm$4.7 & 92.8$\pm$1.3 & 55.4$\pm$5.0 
& 59.3$\pm$1.6 & - & 48.7$\pm$2.3 
& 24.0$\pm$2.8 & 89.2$\pm$3.2 & 18.9$\pm$2.8 
& 65.6$\pm$2.4 & 94.3$\pm$1.1 & 53.5$\pm$3.4 
& 40.2$\pm$5.2 & 79.7$\pm$3.8 & 36.6$\pm$4.9 
& 49.8$\pm$3.3 & 89.0$\pm$1.7 & 42.6$\pm$3.7\\

Qwen3-VL-4B-Thinking
& 62.0$\pm$2.7 & 84.3$\pm$2.2 & 59.5$\pm$2.3 
& 60.0$\pm$2.5 & - & 47.1$\pm$2.3 
& 36.7$\pm$1.2 & 80.2$\pm$1.0 & 31.3$\pm$1.4 
& 52.9$\pm$1.8 & 84.3$\pm$1.0 & 43.2$\pm$2.0 
& 27.8$\pm$1.2 & 78.4$\pm$2.0 & 27.7$\pm$1.3 
& 47.9$\pm$1.9 & 81.8$\pm$0.9 & 41.8$\pm$1.9 \\

Qwen3-VL-8B-Instruct
& 64.3$\pm$1.1 & 89.5$\pm$1.3 & 48.9$\pm$1.1 
& 63.3$\pm$1.9 & - & 46.0$\pm$2.0 
& 34.2$\pm$1.2 & 69.3$\pm$0.9 & 27.2$\pm$1.9 
& 53.1$\pm$2.8 & 87.1$\pm$2.4 & 42.9$\pm$2.0 
& 34.9$\pm$2.6 & 78.5$\pm$1.6 & 30.1$\pm$2.3 
& 49.9$\pm$1.9 & 81.1$\pm$0.9 & 39.0$\pm$1.9  \\

Qwen3-VL-8B-Thinking
& 62.6$\pm$3.3 & 86.9$\pm$1.5 & 57.5$\pm$3.7 
& 61.4$\pm$1.4 & - & 45.8$\pm$1.0 
& 33.3$\pm$1.6 & 84.1$\pm$3.3 & 33.2$\pm$1.7 
& 47.6$\pm$2.0 & 82.0$\pm$2.1 & 40.6$\pm$1.9 
& 43.3$\pm$2.0 & 82.7$\pm$1.5 & 33.8$\pm$2.0 
& 49.6$\pm$2.1 & 83.9$\pm$0.8 & 42.2$\pm$2.1\\

\midrule

\multicolumn{19}{l}{\textbf{Our Model}} \\

\textbf{SenseHD}
& 92.1$\pm$1.8 & 95.3$\pm$1.6 & 92.0$\pm$1.7
& \textbf{97.2$\pm$1.8} & - & \textbf{97.5$\pm$1.6} 
& \textbf{90.8$\pm$1.3} & \textbf{100.0$\pm$0.0} & \textbf{90.8$\pm$1.3} 
& 84.0$\pm$0.9 & \textbf{100.0$\pm$0.0} & 80.6$\pm$2.6 
& 99.9$\pm$0.2 & \textbf{100.0$\pm$0.0} & 99.9$\pm$0.2 
& \textbf{92.8$\pm$1.2} & \textbf{98.8$\pm$0.4} & \textbf{92.2$\pm$1.5}\\
\bottomrule
\end{tabular}
}
\end{table*}

\subsection{HDC-Based Supervised Learning}
\label{subsec:hdc_learning}

\begin{figure}[]
  \centering
  \includegraphics[width=0.5\textwidth]{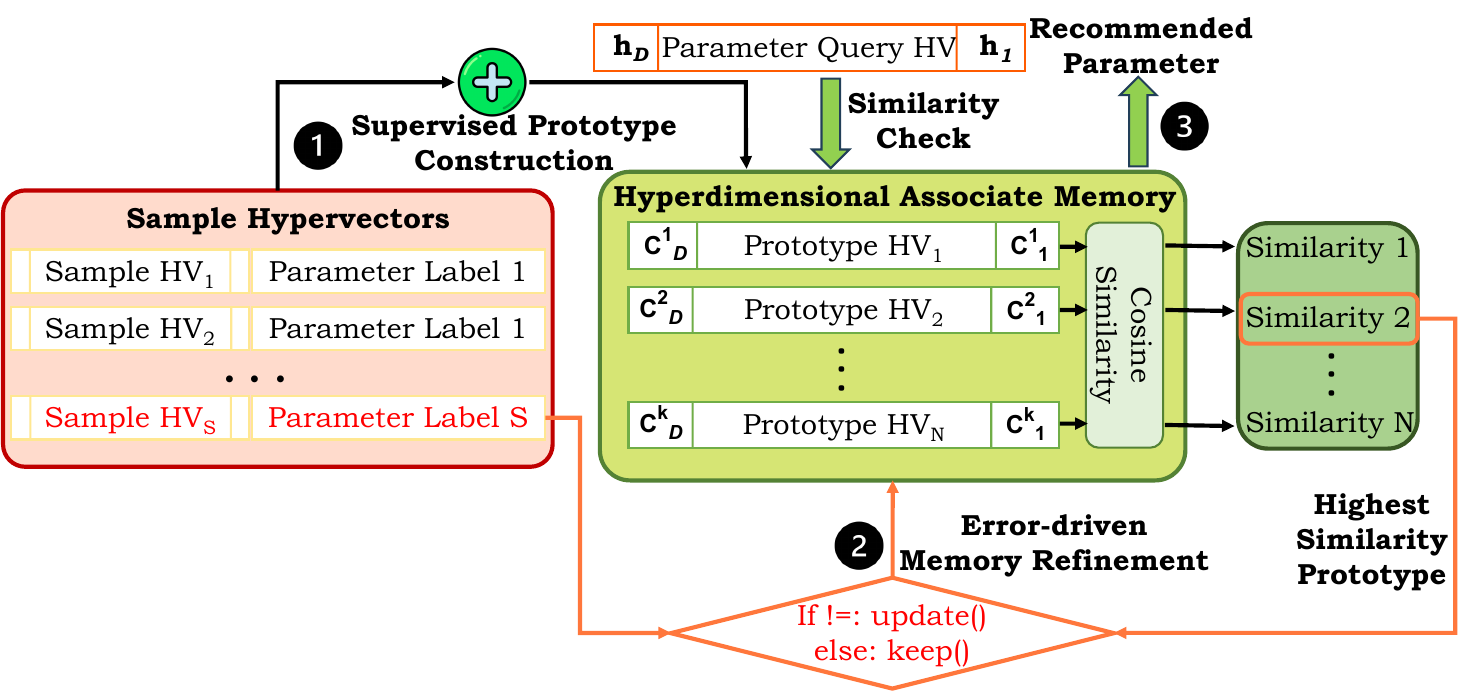} 
  \caption{HDC-based supervised learning of SenseHD.}
  \label{fig:hdc_learning}
\end{figure}

Given the encoded sample hypervectors $\mathbf{h}$, SenseHD formulates scanner parameter learning as a supervised learning problem in hyperdimensional space.
As illustrated in Fig.~\ref{fig:hdc_learning}, this stage focuses on constructing parameter-specific associative memories through supervised prototype learning (\ding{182}) and optional error-driven refinement (\ding{183}).

\paragraph{Supervised Prototype Construction (\ding{182}).}
For each scanner parameter, SenseHD maintains an independent hyperdimensional associate memory composed of a small number of prototype hypervectors, one for each discrete parameter level.
Given a labeled training set $\{(\mathbf{h}_i, y_i)\}_{i=1}^{S}$, where $\mathbf{h}_i \in \mathbb{R}^{D}$ denotes a sample hypervector and $y_i \in \{1,\dots,N\}$ is the corresponding parameter label, samples sharing the same label are aggregated to form a class prototype.
Specifically, for class $k$, the prototype hypervector $\mathbf{c}_k \in \mathbb{R}^{D}$ is obtained by bundling the associated samples,
\begin{equation}
\mathbf{c}_k = \mathcal{A}\big(\{\mathbf{h}_i \mid y_i = k\}\big),
\end{equation}
where $\mathcal{A}(\cdot)$ denotes the HDC aggregation operation.
This prototype construction can be completed in a single pass over the training data.

\paragraph{Error-driven Memory Refinement (\ding{183}).}
Although single-pass prototype aggregation enables fast and efficient training, industrial inspection data may contain ambiguous samples and closely spaced parameter levels that benefit from additional refinement.
To address this, SenseHD adopts an error-driven memory refinement strategy inspired by OnlineHD\cite{hernandez2021onlinehd}. Given a training sample hypervector $\mathbf{h}_i$ with ground-truth parameter label $y_i$, the associative memory first computes the similarity between $\mathbf{h}_i$ and all class prototypes.
Let $\mathbf{c}_{y_i}$ denote the prototype corresponding to the correct label, and $\mathbf{c}_{\hat{y}_i}$ denote the prototype with the highest similarity when a misprediction occurs, where $\hat{y}_i \neq y_i$.
If the prediction is correct, the memory remains unchanged; otherwise, the prototypes are updated according to
\begin{equation}
\mathbf{c}_{y_i} \leftarrow \mathbf{c}_{y_i} + \eta \, (1 - \delta_{y_i}) \, \mathbf{h}_i,
\qquad
\mathbf{c}_{\hat{y}_i} \leftarrow \mathbf{c}_{\hat{y}_i} - \eta \, (1 - \delta_{\hat{y}_i}) \, \mathbf{h}_i,
\end{equation}
where $\eta$ is a learning rate and $\delta_{y_i}$ and $\delta_{\hat{y}_i}$ denote the similarity between $\mathbf{h}_i$ and the correct and mispredicted prototypes, respectively.
This update rule reinforces the association between a sample and its correct parameter class while suppressing interference from competing prototypes.

\subsection{Deployment for Scanner Parameter Recommendation via HDC Inference}
\label{sec:deployment}

SenseHD is deployed as a lightweight inference module for real-time scanner parameter recommendation (\ding{185}).
At deployment time, the learned hyperdimensional associative memory is fixed, and no further parameter updates are performed. Given a new inspection request, the system receives a visual observation and a natural-language instruction describing the inspection intent.
The observation and instruction are encoded into a query hypervector $\mathbf{h}_q \in \mathbb{R}^D$. For each scanning parameter $p$, SenseHD maintains a parameter-specific associative memory consisting of a small set of class prototypes
$\{\mathbf{c}_p^{(1)}, \mathbf{c}_p^{(2)}, \ldots, \mathbf{c}_p^{(K)}\}$,
where each prototype corresponds to a discrete candidate value in the predefined parameter space.
Inference is performed by computing the similarity between the query hypervector and each prototype using cosine similarity,
\begin{equation}
s_k = \delta(\mathbf{h}_q, \mathbf{c}_p^{(k)}),
\end{equation}
where $\delta(\cdot, \cdot)$ denotes cosine similarity.
The recommended scanner parameter is selected by
\begin{equation}
\hat{k} = \arg\max_k \; \delta(\mathbf{h}_q, \mathbf{c}_p^{(k)}),
\end{equation}
and the corresponding parameter value is returned as the system output.
This process is independently executed for each scanner parameter, enabling modular and interpretable parameter recommendation. By reducing inference to simple hypervector similarity and $\arg\max$ operations, SenseHD achieves low-latency and computationally efficient deployment, making it well suited for real-time industrial inspection.

\section{Experiments}
\label{experiments}

\begin{figure*}[]
  \centering
  \includegraphics[width=\textwidth]{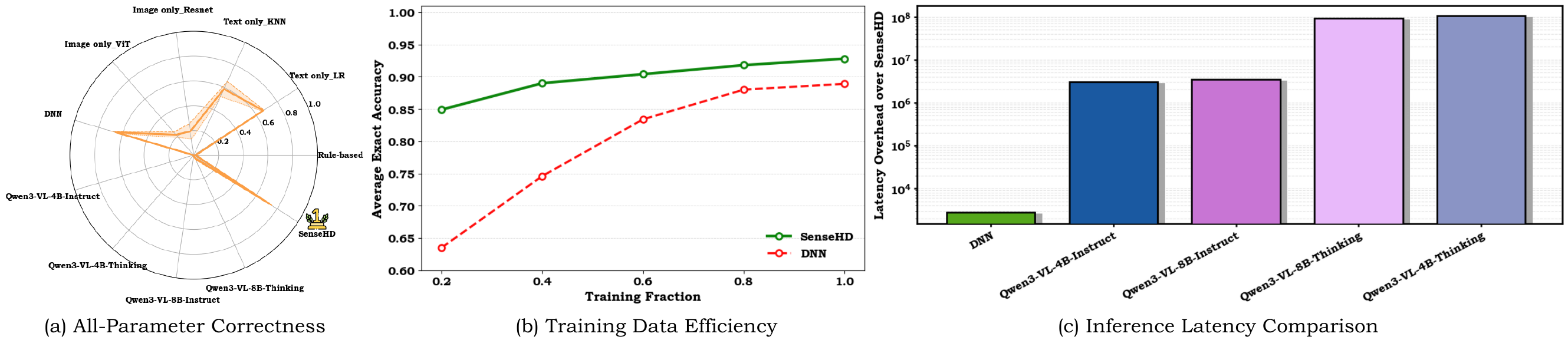} 
  \caption{Quantitative Evaluation of SenseHD.}
  \label{results}
\end{figure*}

\subsection{Experimental Setup and Baselines}

We evaluate SenseHD on the instruction-conditioned scanner parameter recommendation task using the proposed dataset.
All methods are trained and evaluated under identical data splits and evaluation protocols to ensure fair comparison.

\paragraph{Baselines}
We compare SenseHD against a comprehensive set of baselines designed to isolate the contributions of instruction semantics, visual observations, and their multimodal integration, as well as state-of-the-art multimodal large language models (MLLMs).
Rule-based baselines follow manually designed heuristics for scanner parameter selection.
Instruction-only baselines use natural-language instructions as the sole input; we evaluate Logistic Regression and a k-nearest neighbor(KNN) classifier operating on CLIP text embeddings to assess how much parameter information can be inferred from inspection intent alone.
Observation-only baselines rely exclusively on visual input, using ResNet \cite{he2016deep} and ViT \cite{dosovitskiy2020image} models trained on RGB observations to measure the contribution of appearance cues.
As a conventional multimodal learning baseline, a Deep Neural Network (DNN) fuses VLM-extracted visual and textual features via concatenation followed by a three-layer MLP.
We further compare against multimodal large language models that directly predict scanner parameters from image–instruction pairs, evaluating Qwen3-VL\cite{Qwen3-VL} models with different parameter scales and reasoning modes (Instruct vs. Thinking).
All multimodal large language models are evaluated in a zero-shot setting without fine-tuning; reasoning-enabled models are allowed to use chain-of-thought prompting but must output discrete parameter choices from the predefined action space.

\paragraph{Evaluation Metrics}
We report three complementary metrics:
\emph{Exact Accuracy}, which measures strict correctness of predicted parameters;
\emph{Win@1 Accuracy}, which allows one-level deviation from the ground truth; it is not reported for Measurement Range X, as adjacent levels correspond to substantially different fields of view.
and \emph{Macro-F1}, which evaluates class-balanced performance across parameter categories.
All results are reported as mean $\pm$ standard deviation over ten runs.

\paragraph{Implementation Details}

All experiments are conducted on a single NVIDIA RTX~4090 GPU.
For our HDC-based model, we set the hypervector dimensionality to 1000.
The dataset is randomly split into training and test sets with an 8:2 ratio.
For stochastic learning methods, we repeat each experiment multiple times with different random seeds and report the average performance.

\subsection{SenseHD Model Evaluation}

Table~\ref{tab:main_results} summarizes the benchmark results for predicting five scanning parameters. SenseHD consistently outperforms all baselines on Exact Accuracy, Win@1, and F1, achieving the best average performance across parameters.
Rule-based heuristics perform poorly (average Exact Accuracy $<40\%$), indicating that manual rules are insufficient for robust configuration.
Instruction-only and observation-only baselines achieve only partial success: instruction-only models excel on intent-dominated parameters (e.g., 97.9\% Exact Accuracy on sampling frequency) but drop on appearance-sensitive parameters, while observation-only models struggle even on core intent-related parameters (e.g., ResNet at 42.2\% on sampling frequency), showing that visual cues alone cannot capture inspection intent.
The DNN baseline yields moderate gains but remains limited in modeling the structured dependencies needed for reliable reasoning.
Zero-shot multimodal large language models show lower accuracy and higher variance, suggesting that direct generative prediction is unreliable for precise industrial parameter selection.
Beyond per-parameter evaluation, Fig.~\ref{results}(a) adopts a stricter exact-all criterion that requires all five parameters to be predicted correctly.
Under this setting, SenseHD achieves the highest all-parameter correctness, demonstrating superior consistency in joint parameter prediction.
These results validate the effectiveness and robustness of SenseHD for instruction-conditioned scanner parameter recommendation in industrial inspection.

\begin{figure*}[]
  \centering
  \includegraphics[width=\textwidth]{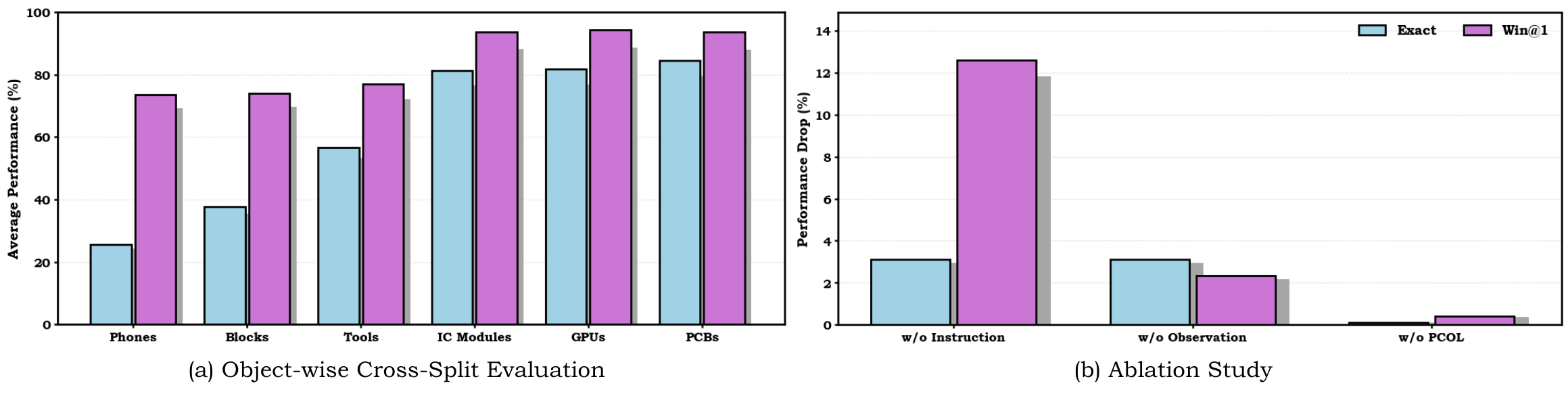} 
  \caption{Object-wise cross-split evaluation and ablation study results.}
  \label{experiment2}
\end{figure*}



\subsection{Training Data Efficiency}

We evaluate the data efficiency of SenseHD against the DNN baseline by varying the training fraction.
As shown in Fig.~\ref{results}(b), SenseHD demonstrates superior few-shot performance, surpassing the DNN by over 20\% when only 20\% of the training data is available.
While the gradient-based DNN struggles to converge with limited supervision, SenseHD's memory-based prototype learning generalizes effectively from sparse samples.
This efficiency is critical for industrial inspection scenarios where collecting large-scale labeled datasets is prohibitive.

\subsection{Efficiency Evaluation}

We evaluate the inference efficiency of SenseHD by comparing its per-sample inference cost with the DNN baseline and multimodal large language models.
The reported latency is measured from receiving a complete sensor sample by the inference pipeline to producing the final prediction.
This measurement includes method-specific preprocessing, encoding, inference computation, and decision generation, while excluding physical sensor acquisition, external communication, logging, and visualization overhead.
As shown in Fig.~\ref{results}(c), SenseHD serves as the zero-overhead baseline, outperforming the DNN baseline which incurs marginal additional latency, while exhibiting substantially lower inference overhead than multimodal large language models.
This efficiency stems from SenseHD’s lightweight inference, which requires only hypervector encoding, similarity computation, and an $\arg\max$ operation, making it suitable for real-time industrial inspection.


\subsection{Object-wise Cross-Split Evaluation}

To evaluate generalization to novel industrial parts, we conduct an object-wise cross-split evaluation where entire object categories are held out during training.
As shown in Figure~\ref{experiment2}(a), SenseHD demonstrates strong generalization on standard electronic components (PCBs, GPUs). For optically challenging objects like Phones (characterized by specular screens and transparency), the Exact Accuracy decreases as expected due to the complex light-matter interactions unique to these materials.
However, the consistently high Win@1 accuracy ($>75\%$ even for Phones) demonstrates that SenseHD successfully learns the underlying physics of scanning.
Instead of random guessing, the model predicts "near-optimal" parameters that serve as a high-quality initialization for new objects. This capability is critical for rapid New Product Introduction (NPI), as it reduces the parameter search space from thousands of combinations to a narrow, feasible window, significantly accelerating the deployment process.


\subsection{Ablation Study}

We further conduct an ablation study to analyze the contribution of instruction conditioning, visual observations, and prompt-conditioned object localization (PCOL). As shown in Fig.~\ref{experiment2} (b), removing instruction conditioning leads to the largest performance drop relative to the full model, with a particularly pronounced degradation in Win@1 accuracy, indicating that task intent is the primary factor constraining predictions to an operationally plausible region of the discrete sensing parameter space. Removing visual observations results in a smaller yet consistent decline in both Exact and Win@1 accuracy, suggesting that visual cues mainly provide appearance-dependent refinement once the task intent is specified. In contrast, ablating PCOL causes only a marginal performance drop, implying that it primarily serves as a robustness-enhancing regularizer rather than a dominant contributor to overall performance.

\subsection{Scan Quality Evaluation}

\begin{figure}[]
  \centering
  \includegraphics[width=0.5\textwidth]{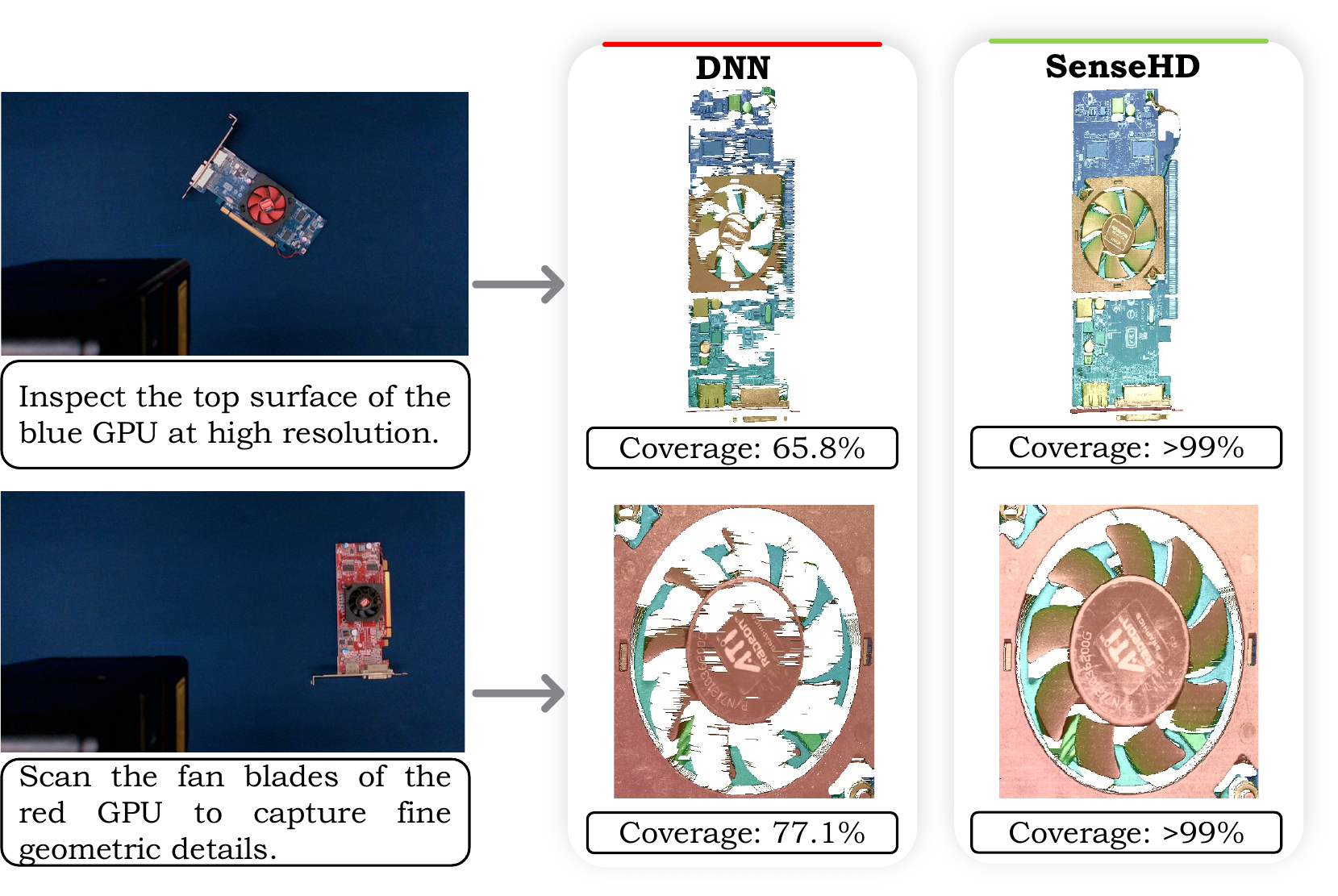} 
  \caption{Comparison of scanning results from different model-selected parameters.}
  \label{quality}
\end{figure}


Beyond offline parameter prediction accuracy, we further evaluate the physical scan quality resulting from model-selected sensing parameters. 
Coverage is computed as the percentage of ground-truth surface points whose nearest-neighbor distance to the reconstructed scan is below a fixed geometric tolerance of 0.1\,mm. 
Fig.~\ref{quality} compares SenseHD with the strongest trained baseline, DNN, under identical RGB observations, inspection instructions, and scanning trajectories, isolating the effect of parameter selection. 
DNN exhibits missing returns and incomplete coverage, achieving 65.8\% coverage for global surface inspection and 77.1\% for fine-grained fan blade scanning. 
In contrast, SenseHD consistently achieves substantially higher coverage under the same strict tolerance across both global and local inspection tasks, producing more stable measurements. 
These results highlight the importance of instruction-conditioned sensing decisions for reliable embodied robotic scanning.

\section{Conclusion}
\label{conclusion}


We present SenseHD, a hyperdimensional memory-centric framework for instruction-conditioned scanner parameter recommendation in robotic inspection. By performing associative retrieval over unified multimodal representations, SenseHD outperforms rule-based, DNN and MLLM baselines on a robotic platform. Results demonstrate high accuracy, data efficiency, and low latency. Future work will extend the framework to richer sensing action spaces and complex inspection scenarios for real-time industrial deployment.





\bibliographystyle{IEEEtran}
\bibliography{refs}


\addtolength{\textheight}{-12cm}   

\end{document}